\documentclass[conference,a4paper]{APSIPA2021}
\usepackage{amsmath}
\usepackage{graphicx}
\usepackage{subfigure}
\usepackage{multirow}
\usepackage{threeparttable}
\usepackage[backend=biber,style=ieee,]{biblatex}
\usepackage{geometry}
\usepackage{fancyhdr}

\fancypagestyle{firststyle}{
  \fancyhf{}
  \fancyhead[C]{2024 Asia Pacific Signal and Information Processing Association Annual Summit and Conference (APSIPA ASC)}
}

\begin{document}

\title{LoFLAT: Local Feature Matching using Focused Linear Attention Transformer}

\author{
\authorblockN{Naijian Cao, Renjie He$^*$, Yuchao Dai, and Mingyi He}
\authorblockA{
Northwestern Polytechnical University, Xian, China\\
E-mail: caonaijian@mail.nwpu.edu.cn, \{davidhrj; daiyuchao; myhe\}@nwpu.edu.cn }
%
}

\maketitle
\thispagestyle{firststyle}
\pagestyle{fancy}

\begin{abstract}
Local feature matching is an essential technique in image matching and plays a critical role in a wide range of vision-based applications. However, existing Transformer-based detector-free local feature matching methods encounter challenges due to the quadratic computational complexity of attention mechanisms, especially at high resolutions. However, while existing Transformer-based detector-free local feature matching methods have reduced computational costs using linear attention mechanisms, they still struggle to capture detailed local interactions, which affects the accuracy and robustness of precise local correspondences. In order to enhance representations of attention mechanisms while preserving low computational complexity, we propose the LoFLAT, a novel Local Feature matching using Focused Linear Attention Transformer in this paper. Our LoFLAT consists of three main modules: the Feature Extraction Module, the Feature Transformer Module, and the Matching Module. Specifically, the Feature Extraction Module firstly uses ResNet and a Feature Pyramid Network to extract hierarchical features. The Feature Transformer Module further employs the Focused Linear Attention to refine attention distribution with a focused mapping function and to enhance feature diversity with a depth-wise convolution. Finally, the Matching Module predicts accurate and robust matches through a coarse-to-fine strategy. Extensive experimental evaluations demonstrate that the proposed LoFLAT outperforms the LoFTR method in terms of both efficiency and accuracy. 
\end{abstract}

\section{Introduction}
\renewcommand{\thefootnote}{}
\footnotetext{This work is partially supported by National Key Research and Development Project of China (2023YFF0906203), Shaanxi Key R\&D Program (2023-ZDLGY-46)\\ * denotes the corresponding author\\}
Image matching is a fundamental task in computer vision, and is crucial for various applications including object recognition~\cite{hou2023survey}, 3D reconstruction~\cite{schonberger2016structure,rao2020nlca}, and autonomous navigation~\cite{zhang2022learning}. 
This process entails the alignment and comparison of images to accurately identify corresponding features or regions, which is essential for achieving precise visual correspondence across varying viewpoints, scales, and lighting conditions.
Among the various techniques employed for image matching, local feature matching stands out as a key approach. 
This technique focuses on the identification and comparison of distinctive local features within images, effectively managing variations in viewpoint, scale, and illumination by leveraging local descriptors that encapsulate unique aspects of the image content. 
Generally, existing local  feature matching methods can be broadly classified into two categories: detector-based and detector-free approaches~\cite{ma2021image}.

Detector-based methods for local feature matching are grounded in the use of specific algorithms to identify salient points in images, such as corners or edges, which are robust to various transformations including scale, rotation, and changes in illumination. 
After these key points are detected, feature descriptors are computed to encapsulate the local image information around each point, facilitating the matching process by comparing these descriptors between different images. 
Methods such as SIFT~\cite{lowe2004distinctive}, SURF~\cite{bay2006surf}, and ORB~\cite{rublee2011orb} exemplify this category, offering high robustness and accuracy in identifying correspondences in structured and well-textured environments. However, these methods have notable limitations, particularly in scenarios with repetitive patterns, texture-less regions, or significant viewpoint changes, where the reliance on distinct key points may lead to inadequate matches and decreased performance.

\begin{figure*}[h]
\begin{center}
\includegraphics[width=0.9\textwidth]{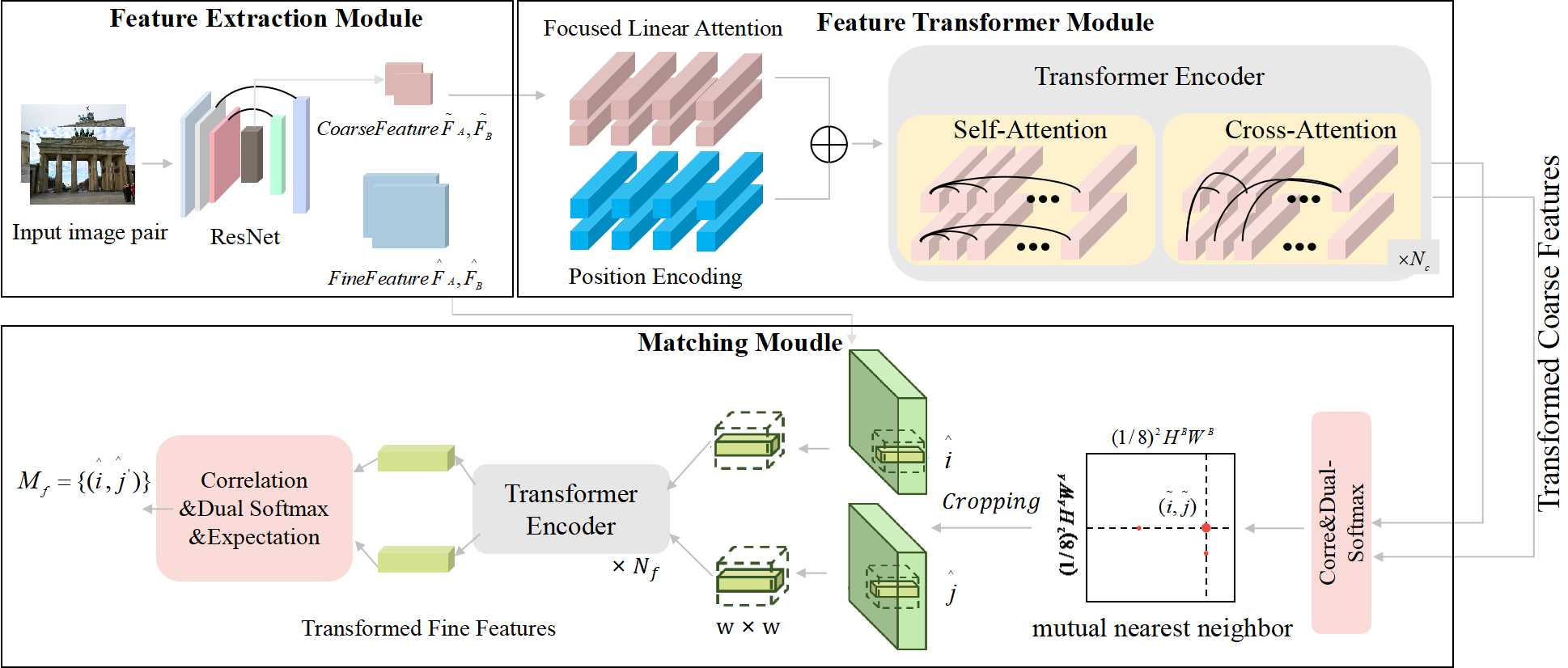}
\end{center}
\vspace*{-6pt}
\caption{Overview of the proposed LoFLAT.}
\label{fig:frame}
\end{figure*}

On the other hand, detector-free methods, also known as dense or end-to-end learning-based methods, avoid the explicit detection of key points~\cite{li2020dual}. 
Instead, they focus on matching features across all pixels or dense grids within images, thereby leveraging the full image content. 
These methods utilize deep learning architectures, such as Convolutional Neural Networks (CNNs) and Transformers~\cite{vaswani2017attention}, to jointly learn feature representations and matching functions. 
By training on large datasets, these approaches can effectively handle a wide range of challenging conditions, including texture-less regions, complex lighting, and severe geometric transformations. 
Techniques like FlowNet~\cite{dosovitskiy2015flownet} and SuperGlue~\cite{sarlin2020superglue} exemplify this paradigm, offering superior flexibility and adaptability in diverse visual tasks.
More recently, methods based on the Transformer backbone have emerged to better model long-range dependencies~\cite{jiang2021cotr,chen2022aspanformer,sun2021loftr,wang2024efficient}. 
Through self-attention mechanisms, Transformers can simultaneously consider all pixels or feature points in an image, enabling feature matching within a global context. 
This capability is particularly beneficial  for addressing  significant  viewpoint changes or complex scenes. 
As one representative work, LoFTR~\cite{sun2021loftr} employs self-attention and cross-attention blocks to update cross-view features. 
Notably, LoFTR integrates a linear Transformer to replace the global full attention mechanism, thereby achieving a manageable computational cost.
It also employs self-attention and cross-attention blocks to update cross-view features.
However, studies have shown that the cross-attention maps generated by the linear Transformer in LoFTR tend to spread over larger areas rather than concentrate on the actual corresponding areas. 
Consequently, LoFTR encounters a significant limitation in extracting highly accurate and stable local correspondences due to the deficiency in detailed local interactions between pixel labels.

To address this issue, this paper introduces a novel local feature matching method based on Focused Linear Attention Transformer. 
The proposed LoFLAT consists of three key modules: the feature extraction module (FEM), the feature Transformer module (FTM), and the matching module (MM).
Specifically, the FEM firstly employs ResNet and a Feature Pyramid Network (FPN) to extract multi-level features from the two images to be matched. 
After that, the FTM is applied to further capture contextually relevant and position-dependent local features.
Specifically, a focused mapping function is utilized to ensure a sharp attention distribution by adjusting the directions of each query and key feature while simultaneously repelling dissimilar pairs.
 Moreover, depth-wise convolution is incorporated  to enhance feature diversity and capture the deep contextual structure of the input images.
Finally, the MM predicts accurate and robust feature matches using a coarse-to-fine framework.
Experimental results on the MegaDepth dataset demonstrate that the proposed LoFLAT outperforms LoFTR in terms of efficiency and feature representation quality.

\section{Proposed Method}
As shown in Fig.~\ref{fig:frame}, the feature extraction module first employs ResNet~\cite{he2016deep} as the backbone network for feature extraction. 
In order to extract features at different resolutions, a feature pyramid network (FPN)~\cite{lin2017feature} is also incorporated to construct a multi-scale feature pyramid. 
After that, features are transformed by the feature transformer module, which leverages self-attention and cross-attention layers to capture local features that are both contextually relevant and position-dependent.
Finally, the matching module outputs precise and robust feature matches through a coarse-to-fine process.

\subsection{Feature Extraction Module}
As illustrated in Fig.~\ref{fig:frame}, the FEM utilizes the ResNet with a feature pyramid network to extract multi-scale feature maps.
In our work, coarse-level features $\tilde{F}_A, \tilde{F}_B$ are extracted with $\frac{1}{8}$ of the original image size, and fine-level features $\hat{F}_B, \hat{F}_B$ are obtained with $\frac{1}{2}$ of the original image size, respectively.

\subsection{Feature Transformer Module}
After the extraction of local features, the Feature Transformer module further is subsequently applied to transform the initial features $ \tilde{F}_A $ and $ \tilde{F}_B $ into $ \tilde{F}_{A}^{t} $ and $ \tilde{F}_{B}^{t} $ for improved representation. 
The objective of this module is to integrate positional and contextual information, thereby augmenting the robustness and discriminative power of the features during the matching process. 
The transformed feature representations  $ \tilde{F}_{A}^{t} $ and $ \tilde{F}_{B}^{t} $ facilitate easier matching, significantly improving both the accuracy and robustness of the matching process. 
It considers not only the intrinsic properties of local features but also the inter-feature relationships and global contextual information, thereby providing a more holistic capture of the structural details within the images.

The core component of Vision Transformer (ViT) is the Transformer encoder, which comprises multiple self-attention layers and feed-forward neural network layers designed to capture global contextual information.  
By employing the self-attention mechanism, ViT is able to aggregate and interact with information across the entire image, thus extending beyond the constraints of a local receptive field. 
Typically, these attention layers employ either standard Softmax attention or Linear attention.

\subsubsection{Softmax Attention}
Softmax Attention is the most basic form of attention in transformer, which assigns weights to value elements by computing the similarity between queries and keys, generating output vectors that allow the model to process information in parallel and capture global context.
The Softmax Attention function can be defined as follows:
\begin{equation}
  \operatorname{Att}(Q, K, V)=\operatorname{softmax}\left(Q K^T\right) V,
\end{equation}
where $Q$, $K$, and $V$ refer to the  query, key and  value vectors, respectively.
However, in high-resolution image feature matching tasks, the number of feature points significantly increases, leading to a substantial growth in token size. 
This makes the employment of Softmax Attention impractical due to the computational demands.

\subsubsection{Linear Attention}
In practice, directly employing Softmax attention becomes challenging and often results in prohibitively high computational costs due to the quadratic complexity of Softmax attention with a global receptive field.
As a result, most approaches primarily employ Linear attention as an efficient alternative due to its superior computational efficiency, lower memory requirements, and effectiveness in real-time processing.
In contrast to the $\mathcal{O}\left({N}^{2}\right)$ complexity of Softmax attention, Linear attention reduces the complexity to $\mathcal{O}\left(N\right)$, significantly enhancing the efficiency of processing large-scale image data. 
This makes Linear attention particularly suitable for resource-constrained environments and applications with stringent real-time requirements. 
The calculation of Linear attention is expressed as follows:
\begin{equation}
\operatorname{Att_L}(Q, K, V)=\phi(Q)\phi(K)^T V,
\end{equation}
where $\phi(x)$ denotes the kernel function that maps data into a new space through nonlinear transformations, while also structurally altering the computation order. 
Linear attention constrains the model to consider only a small subset of key points most relevant to each query point, which reduces the number of necessary comparisons and mitigates the computational bottleneck faced by traditional attention mechanisms when dealing with large-scale datasets.

\begin{figure}[th]
  \centering
        \includegraphics[width=1\linewidth]{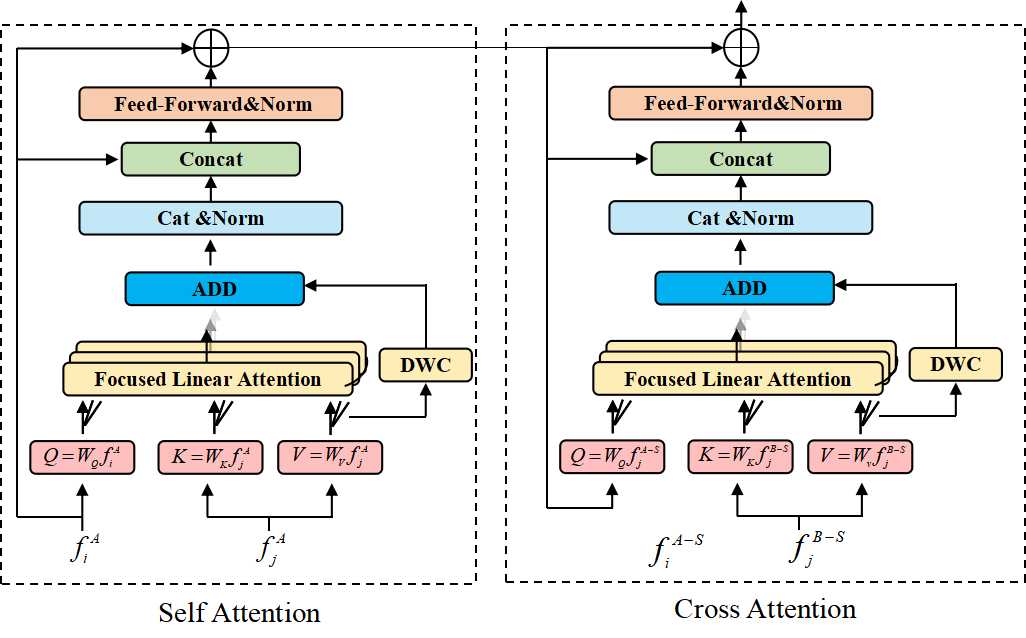}
        \vspace{-10pt}
      \caption{Architecture of the Feature Transformer Module. }
  \label{fig:FLA}
\end{figure}

However, existing Linear Attention methods still require a balance between model complexity and representational capacity. 
While using simple approximations, such as ReLU and ELU activations, may lead to a decrease in performance, more complex kernel function designs or matrix decomposition techniques can result in increased computational costs. 
Furthermore, Linear Attention may be insufficient in capturing long-range dependencies, which can negatively impact the accuracy of feature matching.

\subsubsection{Focused Linear Attention based Transformer}

Previous studies have demonstrated that Linear attention exhibits a relatively smoother distribution compared with Softmax attention. 
This suggests that the attention weights in Linear attention are more uniformly distributed across multiple input features. 
However, for coarse matching, the sharper distribution characteristic of Softmax attention enables the model to focus on a smaller set of key features, thereby enhancing its discriminative capability.
To address both model complexity and representational effectiveness, we propose applying Focused Linear attention~\cite{han2023flatten} in coarse matching.
In the design of the Feature Transformer model, we initially employ a focused mapping function to distill key features from the queries $ Q $ and keys $ K $. 
Following this, we adhere to the principles of Linear Attention to reduce computational cost while maintaining the accuracy of the attention layer output. 
Concurrently, depth-wise convolution is applied to the values $ V $ to capture detailed local features, enhancing the model's sensitivity to fine-grained information.  Ultimately, we integrate the outputs from these two processes to produce a feature representation that balances global context with rich local details. 
The architecture of the Focused Linear Attention based Transformer is illustrated in Fig.~\ref{fig:FLA}.
This approach not only significantly enhances the performance of the attention mechanism but also ensures that the model remains computationally efficient while effectively processing information.

Specifically, the Focused Linear attention can be modeled as follows:
\begin{equation}
\operatorname{Att_{FL}}(Q, K, V)=\phi_p(Q) \phi_p(K)^T V+\operatorname{DWConv}(V)
\label{eq:FLA}
\end{equation}

In (\ref{eq:FLA}), the first term incorporates a focused function $\phi_p$ to replace the original kernel function, which is defined as follows:
\begin{equation}
\begin{split}
  \phi_p(x)&=f_p(\operatorname{ReLU}(x)),\\ f_p(x)&=\frac{\|x\|}{\left\|{x}^{* * p}\right\|} x^{* * p},
\end{split}
\end{equation}
where ${x}^{* * p}$ represents element-wise power $p$ of $x$.

The inclusion of the $\operatorname{ReLU}$ function ensures that all input values are non-negative and valid.
This modification is proved to affect only the direction of the feature vectors. 
By adjusting the directions of query and key features, it effectively aligns similar query-key pairs while distancing dissimilar ones. 
This refined adjustment amplifies the focus on pertinent query-key pairs and diminishes attention to irrelevant pairs, thereby resulting in a more pronounced and sharper attention distribution within the Linear Attention framework.

The second term in (\ref{eq:FLA}) is the depth-wise convolution (DWConv), which is employed to further improve the model's capability in capturing local features and increasing feature diversity.
The DWConv component acts as a localized attention mechanism, enabling each query to focus solely on a limited set of features within its spatial neighborhood rather than considering all features globally. 
The benefit of this localized attention is that, even if two queries receive identical attention weights in a traditional Linear Attention mechanism, DWConv ensures that they extract information from distinct local feature subsets. 

\begin{figure*}[th]
  \centering
  \subfigure[]{
      \includegraphics[width=0.42\linewidth]{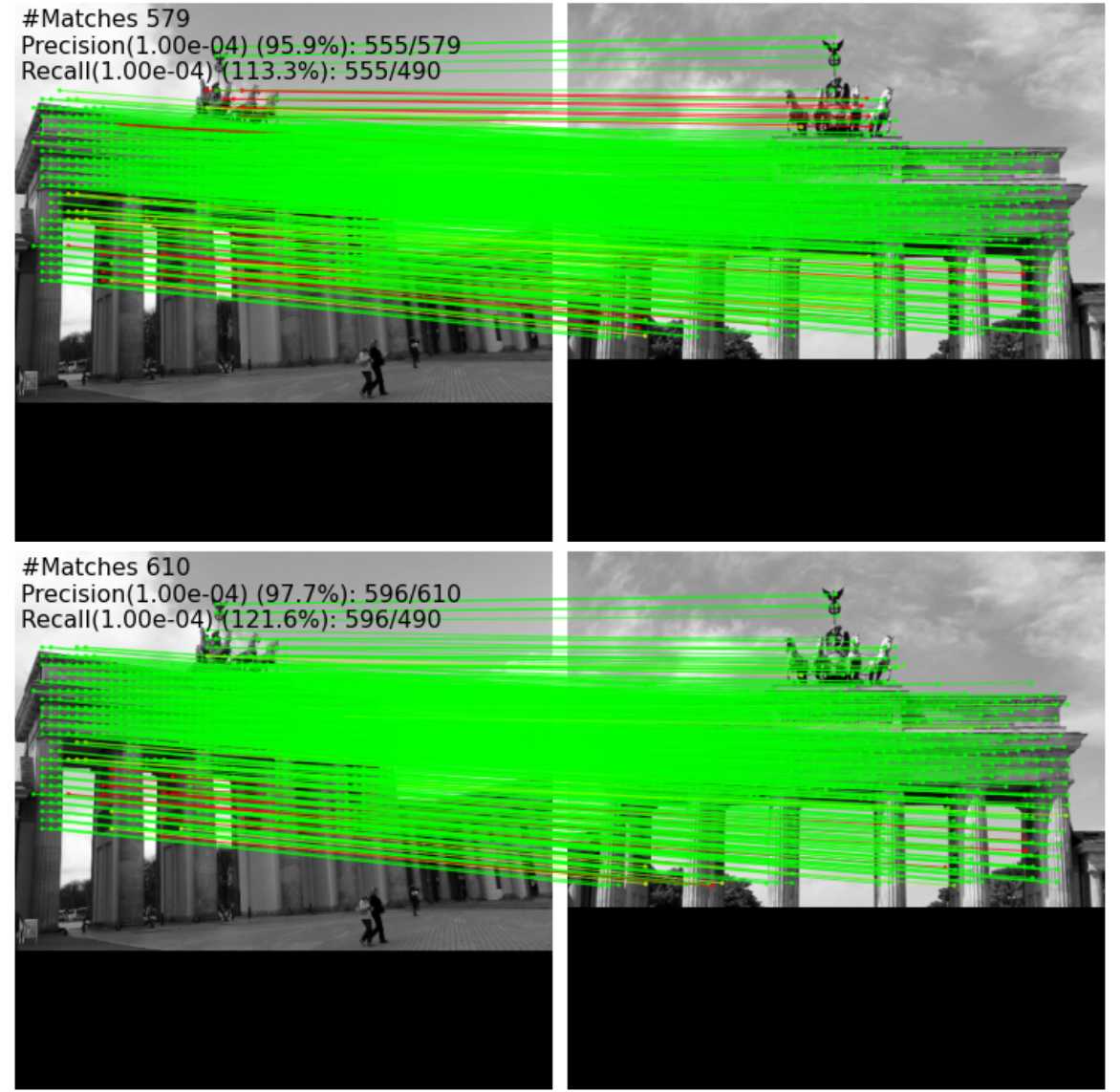}}
  \subfigure[]{      
      \includegraphics[width=0.42\linewidth]{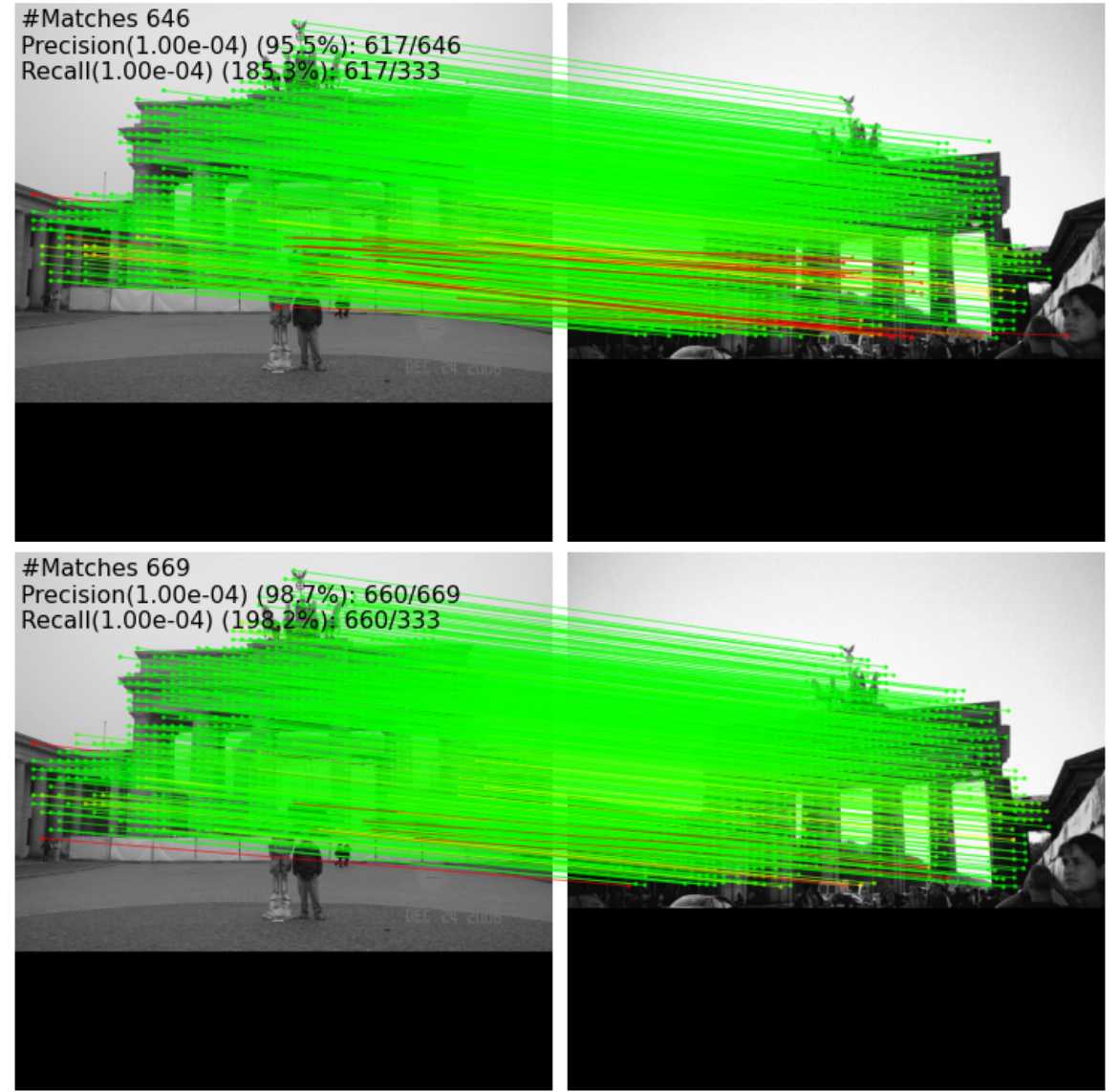}}
  \subfigure[]{
      \includegraphics[width=0.42\linewidth]{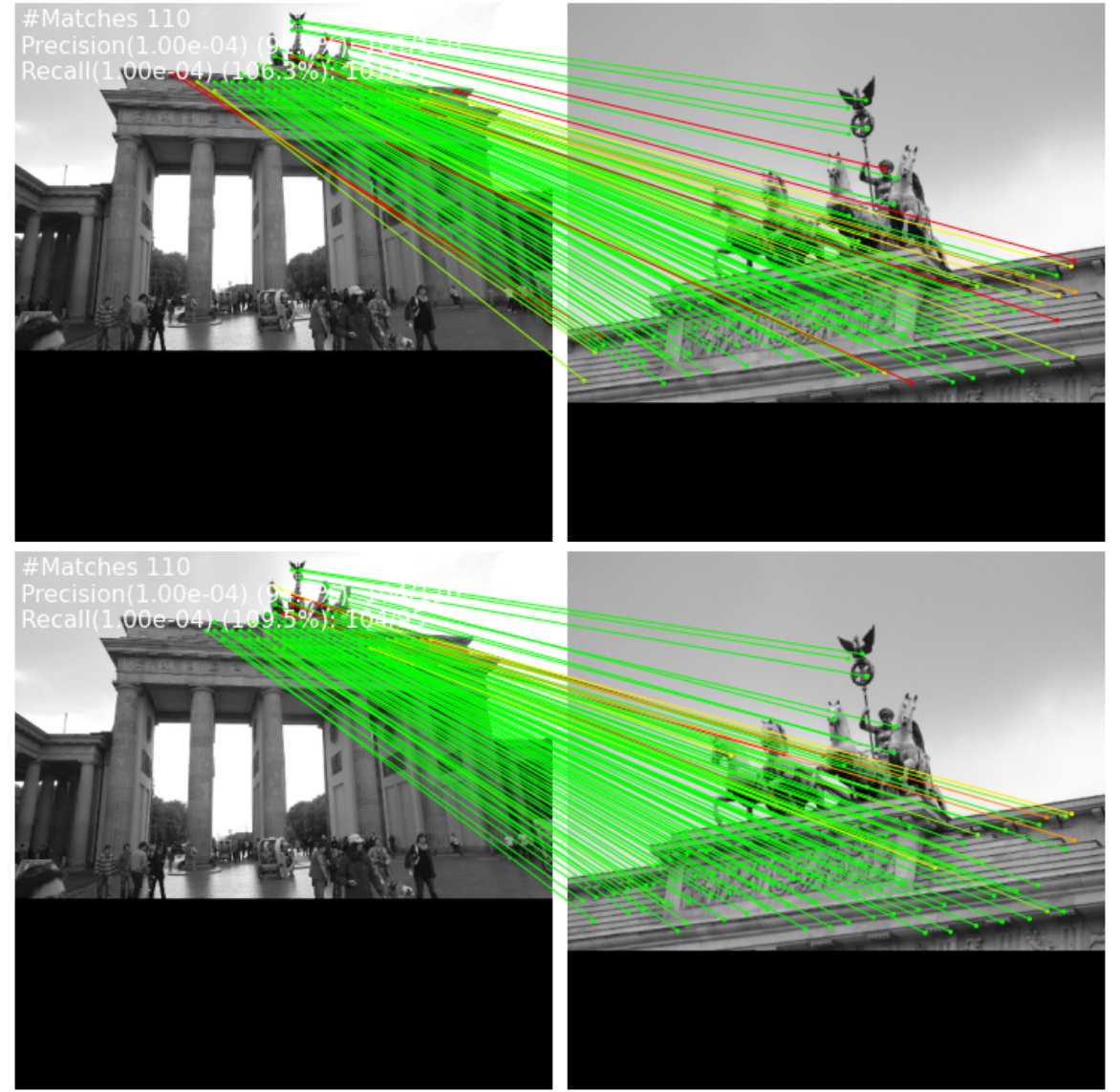}}
  \subfigure[]{
      \includegraphics[width=0.42\linewidth]{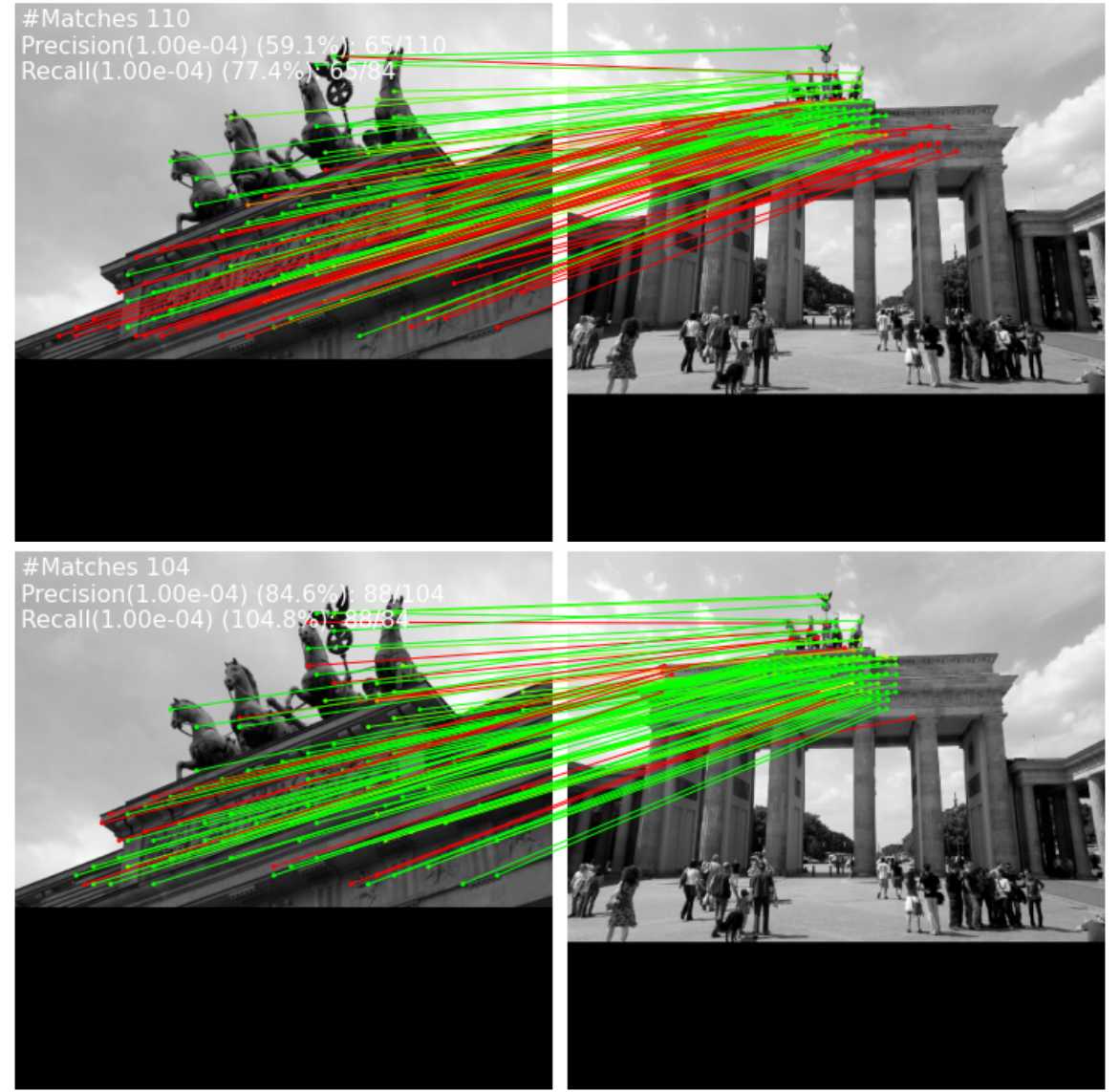}}
    \vspace*{-8pt}
    \caption{Qualitative comparison with LoFTR. top: results of LoFTR, bottom: results of LoFLAT }
  \label{fig:comparison}
\end{figure*}

\subsection{Matching Module}
With both position and context dependent local features obtained, we proceed to establish accurate and robust matching pairs through a  coarse-to-fine  process. 
Specifically, the dual-Softmax operator is firstly applied to determine a coarse level matches by estimating a probability $\mathcal{P}_{coarse}$, which is calculated as follows:
\begin{equation}
  \mathcal{P}_{coarse}(i, j)=\operatorname{softmax}(\mathcal{S}(i, \cdot))_j \cdot \operatorname{softmax}(\mathcal{S}(\cdot, j))_i,
\end{equation}
where $\mathcal{S}$ represents the similarity score matrix, which is calculated by taking the dot product of two feature vectors $\tilde{F}_{\mathrm{t}}^A(i)$ and $\tilde{F}_{\mathrm{t}}^B(j)$ as follows:
\begin{equation}
  \mathcal{S}(i, j)=\frac{1}{\tau} \cdot\left\langle{\tilde{F}}_{t}^A(i), {\tilde{F}}_{t}^B(j)\right\rangle,
\end{equation}
where $\tau$ is a parameter to adjust the smoothness of Softmax operation.


Coarse matches are initially obtained based on the probability matrix $\mathcal{P}_{coarse}$. 
The mutual nearest neighbor criterion is then employed to filter out potential outliers by ensuring bidirectional consistency in the matching pairs.
For each coarse match, the positions are first localized on fine-grained feature maps. 
 Subsequently, local windows are cropped around these localized feature points. 
 A smaller Transformer module is applied to these local feature windows to capture contextual information within the local area.
 Within the local window, the correlation between a feature point and all other points in the same window is calculated, resulting in a heatmap where each pixel value represents the probability of matching with the reference point. 
 By computing the expected value over the probability distribution on the heatmap, the final matching position is obtained with sub-pixel accuracy. 
 Ultimately, the refined positions of each coarse match pair are output with sub-pixel precision, providing a more accurate correspondence between the two images.

\section{Experiments and Analysis}

\subsection{Datasets}
We evaluate the proposed method on the MegaDepth dataset~\cite{li2018megadepth}, which consists of more than one million landmark  photographs from diverse outdoor scenes captured from the This dataset includes a wide range of lighting conditions and scale variations, featuring both day and night images. 
The MegaDepth dataset is used to construct three-dimensional reconstructions of approximately 200 world-renowned landmarks. 
These reconstructions are created using the open-source software COLMAP~\cite{schonberger2016structure}, which processes the images through Structure-from-Motion to derive the camera's internal and external parameters, scene structural information.
The depth maps are then generated through the Multi-View Stereo process. 
For detailed analysis and experimentation, we select tens of thousands high-quality images from the dataset. 
The primary challenge presented by the MegaDepth dataset is achieving accurate matching despite extreme viewpoint changes and repetitive patterns. 
This challenge is addressed by leveraging stereo image pairs derived from the three-dimensional scene information and camera parameters to ensure precise correspondences at the pixel level.

\subsection{Evaluation Metrics}
Following previous work \cite{brachmann2019neural,zhang2019learning,sarlin2020superglue,sun2021loftr}, we report the AUC of the pose error at thresholds of $5^{\circ}, 10^{\circ},$ and $20^{\circ}$ in our experiments. .
The pose error is measured as the maximum of the angular errors in both rotation and translation. 
To determine the camera pose, we use RANSAC to estimate the essential matrix from the predicted matches. 
Moreover, we do not use matching precision for comparison, as there is currently no well-established metric for evaluating detector-free image matching methods.

\subsection{Implementation details}
During the training phase, we utilized 4 NVIDIA GeForce RTX 3090 GPUs, each with a batch size of 1, resulting in a total effective batch size of 4. 
The entire training process spanned approximately 69 hours and consisted of 30 epochs. 
We employed the AdamW optimizer with a weight decay parameter of 0.1. 
The initial learning rate was set to 0.006, with dynamic adjustment of the actual learning rate during runtime. 
To mitigate the risk of gradient explosion, we applied gradient clipping with a threshold of 0.5. 
Additionally, a random seed of 66 was established to ensure reproducibility in data sampling. 
For pose estimation, we also adopted the RANSAC method~\cite{raguram2008comparative}.

The relative pose between image pairs is estimated based on the computed matching positions, and the accuracy of the matching is assessed using AUC. 
Pose error is defined as the maximum value among the angular errors in rotation and translation.

\subsection{Experimental Results}
\thispagestyle{empty}
For fair comparisons, we retrained LoFTR as the baseline and resized the original images to 500 $\times$ 500 pixels.
The comparison of matching between LoFTR and the proposed LoFLAT on MegaDepth dataset is illustrated in Fig.~\ref{fig:comparison}.
 It is observed in Fig.~\ref{fig:comparison} (a) and (b) that both methods are capable of achieving dense matching results in scenarios with small viewpoint changes. However, our method produces a greater number of matches with higher accuracy. 
In Fig.~\ref{fig:comparison} (c) and (d), it is demonstrated that in scenarios with significant viewpoint and scale changes, the accuracy of LoFTR noticeably decreases, leading to an increased number of erroneous matches. 
Our method, on the other hand, significantly improves matching accuracy in these challenging cases.  
Overall, our approach outperforms LoFTR in both match density and accuracy, demonstrating superior robustness and stability.

\thispagestyle{empty}
\begin{table}[]
\centering
\caption{Evaluation of Relative Pose Estimation on MegaDepth Dataset}
\renewcommand\arraystretch{1.2}
\begin{tabular}{lccc}
\hline
            & AUC@5$^{\circ}$ & AUC@10$^{\circ}$ & AUC@20$^{\circ}$ \\ \hline
LoFTR       & 42.9  & 60.6   & 74.8   \\
LoFTR+Focus & 44.9  & 61.7   & 75.1   \\
LoFLAT        & \textbf{45.6}  & \textbf{62.5}   & \textbf{75.9}   \\ \hline
\end{tabular}
\label{tab:compare}
\end{table}

We further compare the performance of LoFTR and our proposed LoFLAT in terms of AUC metric at different angular error thresholds ($5^{\circ}, 10^{\circ},$ and $20^{\circ}$). It is observed in Table~\ref{tab:compare} that our model outperforms the baseline method, LoFTR, across all three thresholds, with improvements of 2.7\%, 1.9\%, and 0.9\%, respectively. 
Notably, the performance gains are more pronounced under stricter threshold conditions (i.e., smaller angular errors). This suggests that our method excels particularly well in tasks requiring precise matching.

\section{Conclusions}
In this paper, we introduced  a novel local feature matching method based on the Focused Linear Attention Transformer, to address the limitations of LoFTR. 
Our method comprises three main modules: the Feature Extraction Module, the Feature Transformer Module, and the Matching Module. 
The Feature Extraction Module leverages ResNet and a Feature Pyramid Network to extract hierarchical features, while the Feature Transformer Module refines attention distribution using a focused mapping function and depth-wise convolution to enhance feature diversity. 
The Matching Module employs a coarse-to-fine strategy to predict accurate and robust matches.
Extensive experimental evaluations demonstrate that the proposed LoFLAT significantly outperforms the LoFTR method in terms of both efficiency and accuracy. 

\printbibliography

\end{document}